# Graph-Based Financial Fraud Detection with Calibrated Risk Scoring and Structural Regularization

Yunfei Nie

Brandeis University, Waltham, USA

Jiawei Wang

University of California, Los Angeles, USA

Ruobing Yan

Georgetown University, Washington, D.C., USA

Yuhan Wang

Columbia University, New York, USA

Zouxiaowei Ma

Columbia University, New York, USA

Yilun Wu*

Stevens Institute of Technology, Hoboken, USA, wuyilun1028@gmail.com

## Abstract

Financial transaction fraud prevention faces challenges such as complex relationship structures, concealed behavioral patterns, and dynamically changing data distribution. Discrimination models relying solely on independent sample features are insufficient to fully characterize the risks of group collaboration and chain transfers within transaction networks. This paper proposes a graph neural network representation learning and risk discrimination framework for financial transaction fraud prevention. It integrates transaction records and identity information into node attributes and constructs a transaction graph based on shared attributes and interaction consistency to explicitly model inter-transaction relationships. In model design, a multi-layer message passing mechanism is employed to aggregate neighborhood information, learn node embedding representations containing structural context semantics, and output transaction-level fraud probability and risk scores through a lightweight risk discrimination head. A weighted supervision objective is introduced to mitigate training bias caused by class imbalance, and structural consistency regularization constraints are combined to suppress the impact of noisy edges on representation drift, thereby improving the stability and usability of risk characterization. Experiments are conducted on a publicly available financial transaction dataset, comparing various methods in the same direction and comprehensively evaluating them under a unified evaluation protocol. The results show that the proposed method outperforms other methods in risk ranking and probability calibration quality, validating the effectiveness of graph structure modeling and representation learning collaboration in financial transaction fraud prevention.

# 1 Introduction

The digitalization and high frequency of financial transactions have driven the rapid expansion of payment, transfer, and online consumption scenarios, but have also made fraud more organized and covert. Fraudsters often evade rule interception through multi-account collaboration, chain-like fund transfers, and cross-scenario masquerading. Risk signals are no longer concentrated on isolated features such as the amount or frequency of a single transaction, but are dispersed across the interaction relationships between accounts, temporal rhythms, and group structures. Simultaneously, the transaction network continues to grow in scale, with more diverse node and relationship types [1]. Data gaps, noisy connections, and strategic adversarial tactics are prevalent, making it difficult for traditional solutions based on static feature engineering and manual rules to maintain stable and effective risk characterization capabilities in the long term [2].

Against the backdrop of increasingly stringent risk control governance and compliance audit requirements, anti-fraud models not only need strong identification capabilities but also need to provide usable risk scores and interpretable clues under high-throughput, low-latency business constraints. In actual business, labels are often lagging and scarce, the proportion of fraud samples is low, and definitions are adjusted with policy and business changes, leading to model training bias and conceptual drift [3]. For risk management, stable probability output and reasonable threshold settings are equally crucial. Overconfidence or unstable fluctuations in the model will directly impact alert strategies, manual review costs, and user experience. Therefore, modeling for financial transaction fraud prevention requires simultaneous attention to the representation of structured relationship information, robustness to imbalances and noise, and the reliability of risk output [4].

Graph representation learning provides a natural modeling paradigm for depicting the relational structure of financial transactions. By mapping transactions and their associated entities to a graph structure, the shared attributes, common environments, and interaction consistency between transactions can be explicitly expressed through edge connections. Patterns such as group collaboration and chain transfers can also be captured during neighborhood propagation and aggregation [5]. Compared to discriminative models that rely solely on independent samples, graph neural networks introduce local structural context into node representations through message passing mechanisms, enabling risk identification to utilize both individual attributes and relationship patterns, thus more closely resembling the generation mechanism of real fraudulent behavior. However, financial transaction graphs typically exhibit sparsity, heterogeneity, and noise contamination. How to suppress the interference of low-quality connections in graph construction and representation learning, and form stable risk scores that can be used for risk control decisions, remains a problem requiring systematic research.

This paper focuses on the above-mentioned needs and makes the following main contributions:

(1) It proposes a transaction graph modeling approach for anti-fraud in financial transactions, unifying multi-source transaction attributes and identity information into node features, and constructing inter-transaction relationships through shared attributes and interaction consistency to support explicit characterization of group structural risks.

(2) It designs a representation learning and risk discrimination framework based on graph neural networks, learning structural context embedding through multi-layer message passing, and outputting transaction-level risk probabilities and risk scores to meet the needs of practical risk control for continuous risk characterization.

(3) It introduces imbalance processing and structural consistency constraints into the optimization objective to improve the training stability and risk output reliability of the model under scarce labels, noisy relationships, and complex network structures.

# 2 Methodology Foundation

The methodological design of the proposed framework is grounded in recent advances in graph-based relational modeling, adaptive learning under distributional uncertainty, and robust risk-oriented optimization strategies. At its core, the approach builds upon graph neural network (GNN) paradigms that explicitly encode inter-transaction dependencies, while incorporating complementary mechanisms from imbalance-aware learning, continual adaptation, and causal reasoning to address the unique challenges of financial fraud detection.

A primary foundation of the proposed method lies in multi-hop relational modeling over transaction graphs, where dependencies among entities are propagated through structured neighborhoods. Prior work has demonstrated that capturing higher-order interactions significantly enhances fraud detection performance by revealing collaborative and chain-like behaviors that are otherwise invisible in isolated feature spaces [6]. This motivates the use of multi-layer message passing in the current framework, where node representations are iteratively enriched with contextual signals from both immediate and extended neighborhoods. However, unlike existing approaches that primarily emphasize representation expressiveness, the present work further constrains the propagation process through structural regularization to mitigate the amplification of noisy or spurious edges, thereby improving embedding stability in sparse and contaminated financial graphs.

Beyond relational modeling, the construction of meaningful graph structures is informed by knowledge graph-based frameworks that integrate heterogeneous financial entities and relationships into unified semantic representations [7] Such approaches highlight the importance of leveraging shared attributes and contextual consistency to form interpretable connections between transactions. Inspired by this perspective, the proposed method adopts a principled graph construction strategy that combines attribute similarity and interaction coherence, enabling the learned representations to better reflect latent fraud patterns while maintaining interpretability. Nevertheless, existing knowledge-driven methods often rely on predefined schemas or generative reasoning processes, which may limit scalability and real-time applicability. The current framework addresses this limitation by embedding these relational cues directly into a lightweight, end-to-end trainable graph learning pipeline.

The challenge of dynamic and non-stationary transaction environments further necessitates adaptive learning mechanisms. Continual learning frameworks for anomaly detection have demonstrated effectiveness in tracking evolving data distributions through dynamic monitoring and incremental updates [8]. Similarly, distribution-shift-aware learning paradigms emphasize the joint handling of class imbalance and temporal drift to maintain predictive reliability in changing environments [9]. These insights directly inform the design of the weighted supervision objective in this work, where class imbalance is explicitly addressed through adaptive loss weighting. At the same time, the incorporation of structural consistency regularization serves as an implicit stabilization mechanism against distributional perturbations, reducing representation drift without requiring explicit retraining cycles. This design reflects a trade-off between adaptability and computational efficiency, which is critical for high-throughput financial systems.

Another important methodological influence comes from few-shot and meta-learning approaches in fraud detection, which aim to generalize under limited labeled data conditions [10]. These methods highlight the importance of learning transferable patterns and robust decision boundaries when fraud labels are scarce and delayed. While the proposed framework does not explicitly adopt a meta-learning paradigm, it is inspired by the need for label-efficient learning. This is reflected in the emphasis on leveraging structural context as an auxiliary signal to compensate for sparse supervision, thereby improving generalization without increasing labeling requirements.

In addition, causal reasoning frameworks over structured data provide a complementary perspective on disentangling spurious correlations from genuine risk factors [11] [12]. These approaches emphasize intervention-oriented analysis and reasoning over relational structures, which is particularly relevant in fraud detection where observed patterns may be influenced by confounding behaviors or adversarial manipulation. Although the proposed method does not explicitly model causal graphs, the introduction of structural consistency constraints can be interpreted as a proxy for enforcing invariant relationships across connected nodes. By encouraging neighboring transactions with consistent relational semantics to exhibit coherent representations, the model implicitly reduces the influence of noisy or non-causal connections, thereby enhancing the reliability of risk estimation.

Accordingly, the proposed framework operationalizes these methodological insights through a unified design that tightly couples graph-based representation learning with risk-oriented optimization. The model architecture leverages multi-layer message passing to encode structural dependencies, while the training objective integrates imbalance-aware weighting and structural consistency constraints to jointly address label sparsity, noisy connectivity, and distributional instability. This formulation enables the learned representations to remain both discriminative and stable under realistic financial environments, and ensures that the resulting risk probabilities exhibit improved calibration and usability for downstream decision-making.

# 3 Methodology

## 3.1 *Dataset*

The experiment utilizes the IEEE CIS Fraud Detection open-source dataset, publicly released by Kaggle. This dataset is designed for large-scale online transaction fraud detection tasks, providing transaction records and anonymized features in a real-world business context. It can be used to build transaction risk discrimination models and evaluate generalization capabilities. The data uses the binary label 'isFraud' to indicate whether a transaction is fraudulent. The dataset contains approximately 590,000 transactions, including a mixture of continuous and discrete features, and covers multi-dimensional signals such as device information, email domains, payment, and product-related features. It exhibits typical class imbalance and noise interference characteristics, meeting the robustness and recall sensitivity modeling requirements of financial transaction anti-fraud scenarios.

This dataset consists of two parts: 'transaction' and 'identity', which can be linked via 'TransactionID'. 'transaction' focuses on transaction behavior and attribute features, while 'identity' focuses on identity features related to the network environment and device. Not all transactions possess complete identity information. Based on this structure, each transaction can be treated as a node, and shared attributes such as card number, address, device information, and email domain name can be used to construct the connection edges between transactions, forming a heterogeneous relationship graph to support graph neural network representation learning. At the same time, the original time series fields are retained to characterize dynamic patterns such as short-term aggregation and chain transitions, thereby achieving joint optimization of structural representation and risk judgment under a unified framework.

## 3.2 *Dataset preprocessing*

The original transaction and identity data are first connected based on unique identifiers. Fields with high missing values are then filtered to reduce noise, while retaining the core feature set related to transaction behavior, device environment, and account attributes. For numerical features, missing values are imputed, and outliers are pruned. Robust quantile truncation is used to suppress the interference of extreme amounts and long-tailed distributions on gradient updates; subsequently, standardization is performed to eliminate dimensional differences. Categorical features are uniformly labeled and encoded with missing values. For high-cardinality features, frequency truncation and rare category grouping are used to mitigate dimensionality expansion and overfitting risks. Simultaneously, parsable time fields are converted into periodic representations to characterize intraday and weekly transaction rhythm changes.

Regarding graph structure construction, each transaction is modeled as a node, and relationship edges between transactions are generated based on shared attributes, including consistent connections of payment-related identifiers, addresses, and device environments, to form a network reflecting related transaction groups and potential collaborative behaviors. To improve the effectiveness of structural information, deduplication and threshold filtering are performed on relation edges to limit low-quality connections caused by overly sparse or overly dense attributes. Connectivity component analysis is also performed on the graph to remove isolated, noisy subgraphs. During the training phase, a time-consistent partitioning strategy is adopted to reduce the risk of information leakage. Class imbalance handling is introduced into the loss function and sampling strategy. Through positive and negative sample reweighting and intra-batch negative sampling stabilization optimization process, a clean, controllable, and reproducible input is provided for subsequent graph neural network representation learning and risk discrimination. This paper also presents comparative statistical results before and after dataset preprocessing, as shown in Figure 1.

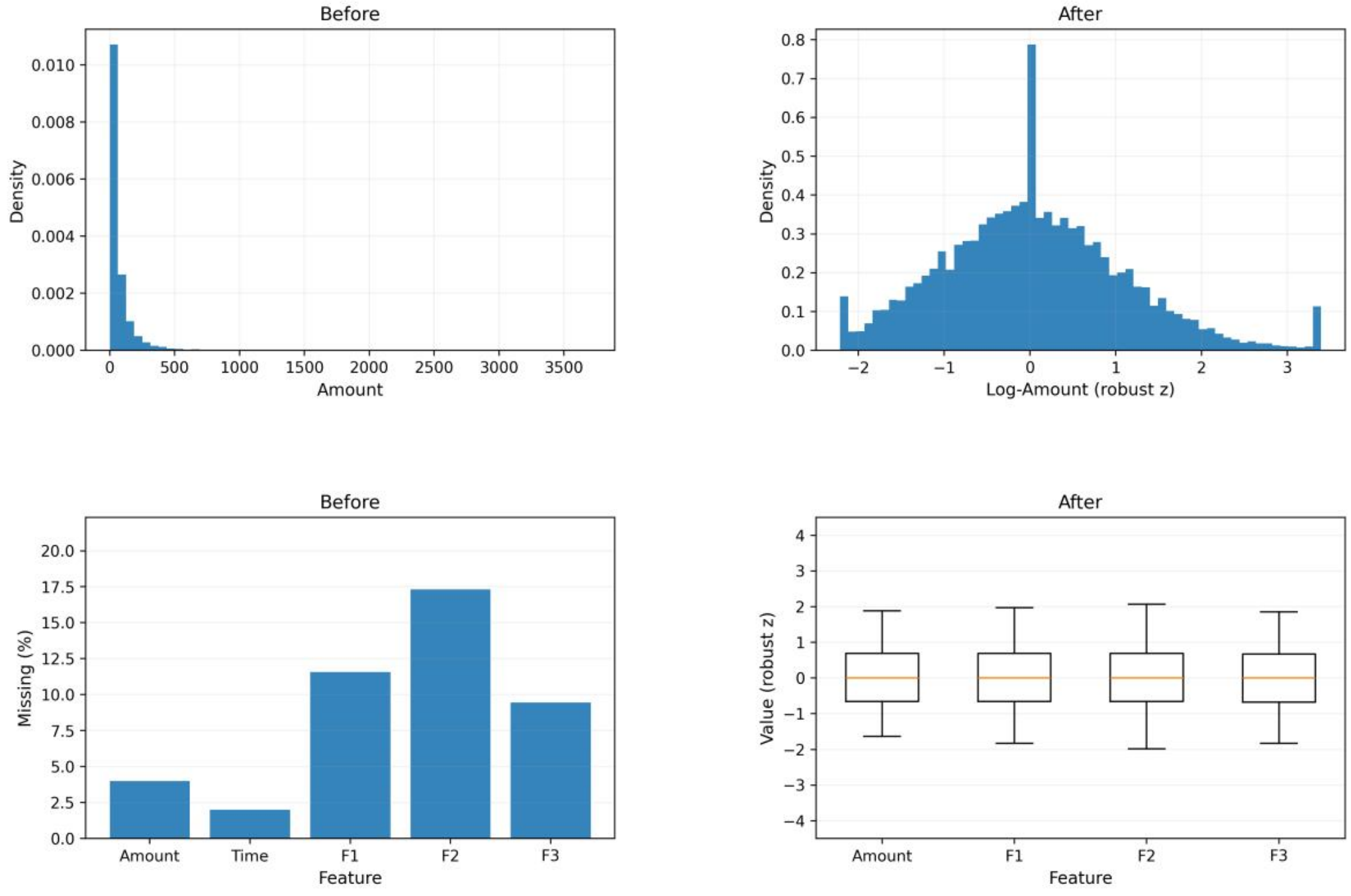

**Figure 1. The figure shows the statistical distribution changes of financial transaction data before and after preprocessing. After preprocessing, the amount distribution is smoother, the missing rate is significantly reduced, and the feature scale is more regular and stable.**

## 3.3 *Overall framework and graph modeling*

The core objective of risk assessment for fraud prevention in financial transactions is to learn stable graph representations under complex transaction relationships and multi-source attribute signals, and to map node-level transaction risks into continuous risk scores that can be used for alerting and handling. The overall process comprises three parts: graph construction, graph neural network representation learning, and risk assessment. First, transaction records and identity-related features are aligned and fused to form an attribute vector for each transaction. Then, relationship edges between transactions are established based on shared attributes and interaction consistency, resulting in a transaction graph capable of characterizing homogeneous transaction groups, short-term aggregations, and chain transitions. Finally, multi-layer message passing is performed through a graph neural network to obtain the structural context representation of each transaction node, and a lightweight discriminator outputs the fraud risk probability. To maintain discriminative stability under noisy and sparse label conditions, the model simultaneously constrains the robustness of neighborhood aggregation and the calibrability of risk output, avoiding risk bias by relying solely on single-point features while ignoring relationship patterns. This paper presents the overall model architecture, as shown in Figure 2.

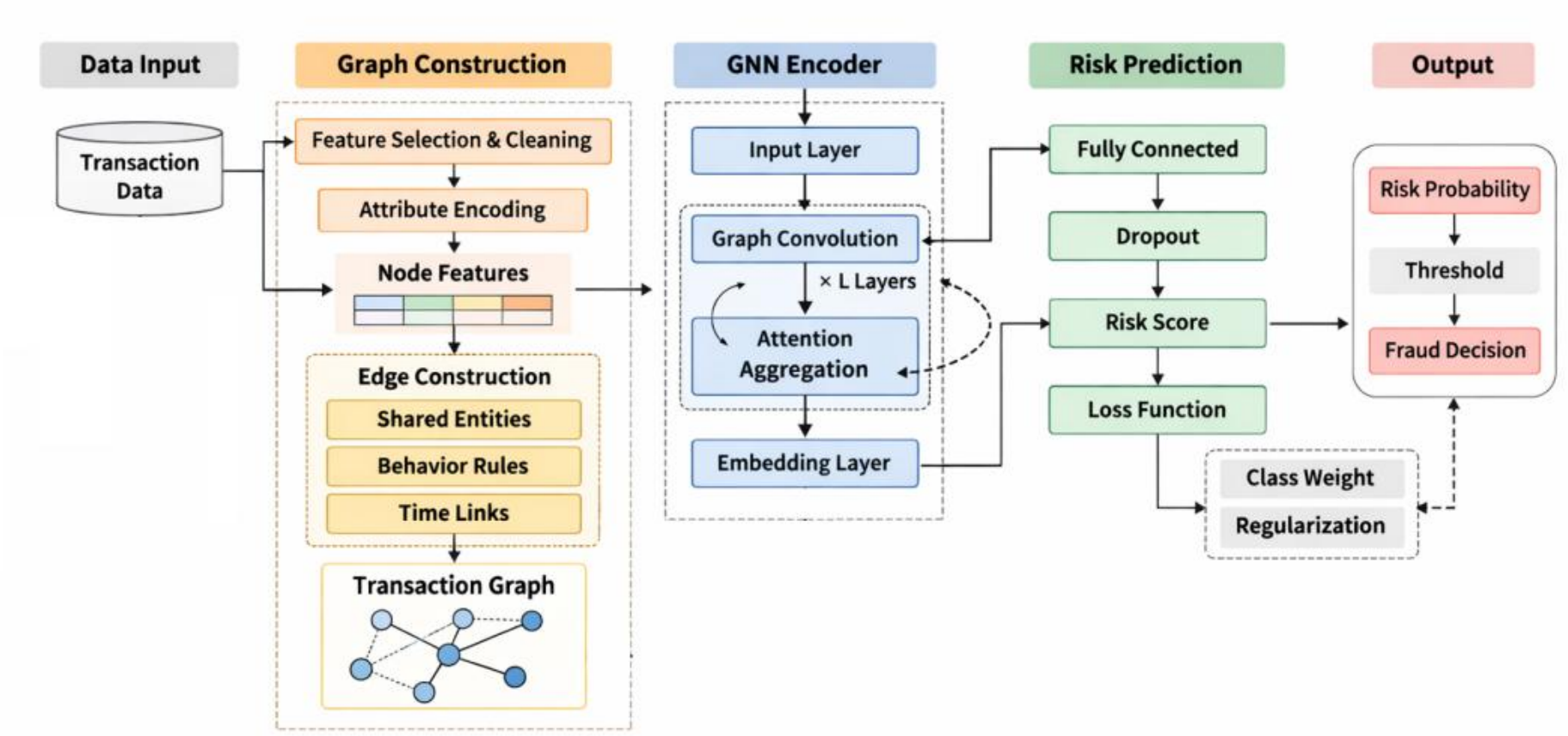

**Figure 2. The overall architecture of the proposed graph representation learning and risk discrimination framework for financial transaction fraud detection covers data fusion, graph construction, GNN-based encoding, and risk prediction. The pipeline outputs transaction-level fraud probability and final decision via a lightweight prediction head with class weighting and regularization.**

The transaction graph is defined as $G=(V,E)$, where the set of nodes $V$ represents transaction samples, and the set of edges $E$ represents the relationships between transactions. Each node $i\in V$ has an initial attribute vector $x_i\in R^d$. Message passing is performed based on the first-order neighborhood $(i,j)\in E$ of the relation edges $N(i)$, using neighborhood aggregation with normalization.

$$h_i^{(0)}=x_i,\ h_i^{(l+1)}=\sigma(W^{(l)}h_i^{(l)}+\sum_{j\in N(i)}\alpha_{ij}U^{(l)}h_i^{(l)}) \qquad (1)$$

Where $\sigma(\bullet)$ is a nonlinear activation function, $W^{(l)},U^{(l)}$ is a learnable parameter, and $\alpha_{ij}$ is an edge weight or attention coefficient, used to control the contribution of different neighbor relationships to the node representation update, and numerical stability and noise suppression are achieved through degree normalization or soft attention.

## 3.4 *Risk identification head and optimization objectives*

After obtaining the node representation $h_i^{(L)}$ following message passing at layer $L$, the risk discrimination module maps it to a transaction fraud probability and outputs a continuous risk score to support thresholding. The node risk probability is obtained using a single-layer linear mapping and Sigmoid activation.

$$\widehat{p}_i=sigmoid(w^Th_i^{(L)}+b) \qquad (2)$$

Where $w$ and $b$ are learnable parameters, and $\widehat{p}_i\in(0,1)$ represents the fraud risk of transaction $i$. To improve the adaptability to class imbalance, a weighted binary cross-entropy is introduced as the main monitoring signal:

$$L_{cls}=-\sum_{i\in V}(\beta y_i log(\widehat{p}_i)+(1-y_i)log(1-\widehat{p}_i)) \qquad (3)$$

Where $y_i\in\{0,1\}$ is the fraud label and $\beta$ is the positive class weight, which is used to enhance the learning strength of scarce fraud samples and reduce the bias dominated by the majority class.

To suppress representation drift caused by noisy relationships and enhance the local consistency of similar transactions in the graph embedding space, an edge-based representation smoothing regularization term is introduced to encourage connected nodes to maintain reasonable proximity in the representation space.

$$L_{smooth} = \sum_{(i,j)\in E} \alpha_{ij} \| h_i^{(L)} - h_j^{(L)} \|_2^2 \qquad (4)$$

The final optimization objective consists of a weighted average of the classification loss and the structure regularization term:

$$L = L_{cls} + \lambda L_{smooth} \qquad (5)$$

Among them, $\lambda$ controls the structural consistency constraint strength, so that risk judgment maintains sensitivity to local relation patterns while avoiding judgment degradation caused by excessive smoothing, thus forming an integrated modeling framework for graph representation learning and risk judgment for anti-fraud in financial transactions.

# 4 Experimental Results and Analysis

## 4.1 *Experimental setup*

The experiment was conducted in a single-machine, single-GPU environment. The software stack used Linux operating system and Python 3.10. The deep learning framework used was PyTorch 2.2, with CUDA 12.1 and cuDNN 8.9 for accelerated training. Model training employed the AdamW optimizer with weight decay to improve generalization stability. The learning rate was decayed using a cosine annealing strategy. The maximum number of training epochs was set to 200, and an early stopping strategy was enabled to avoid overfitting. The batch size was set to 1024 to accommodate large-scale transaction graph sampling training. The graph neural network encoder used a 3-layer message-passing structure with a hidden dimension of 128, 4 attention heads, and Dropout set to 0.30. To address the class imbalance problem, positive class weight coefficients were set, and stratified sampling was performed during training. A fixed random seed was used during training to ensure reproducibility. The specific hardware and software environment and key hyperparameter configurations are summarized in Table 1.

**Table 1. Hardware/Software Environment and Key Hyperparameter Settings**

| Category | Configuration Item | Value |
|---|---|---|
| Hardware | GPU | NVIDIA RTX 3090 (24 GB) |
| Hardware | CPU | Intel Xeon (16 cores) |
| Hardware | Memory | 128 GB |
| Software | Operating System | Ubuntu 22.04 LTS |
| Software | Python | 3.10 |
| Software | Deep Learning Framework | PyTorch 2.2 |
| Software | CUDA | CUDA 12.1 |
| Software | cuDNN | cuDNN 8.9 |
| Training | Optimizer | AdamW |
| Training | Initial Learning Rate | 0.0005 |
| Training | Weight Decay | 0.0001 |
| Training | Learning Rate Scheduler | Cosine Annealing |
| Training | Training Epochs | 200 |
| Training | Batch Size | 1024 |
| Model | Number of GNN Layers | 3 |
| Model | Hidden Dimension | 128 |
| Model | Number of Attention Heads | 4 |
| Model | Dropout | 0.30 |
| Reproducibility | Random Seed | 42 |

## 4.2 *Experimental Results and Analysis*

To verify the effectiveness of the proposed graph neural network representation learning and risk discrimination framework in the anti-fraud task of financial transactions, representative methods in the same research direction were selected for comparison, and their key indicators were summarized under a unified data partitioning and evaluation protocol to conduct a comprehensive evaluation from two dimensions: discrimination ability and risk characterization quality. The specific indicator comparison is shown in Table 2.

**Table 2. Experimental results compared with other models**

| Method | Accuracy | Precision | Recall | F1 | AUROC | AUPRC | ECE | Brier Score |
|---|---|---|---|---|---|---|---|---|
| Cheng et al.[13] | 0.87 | 0.86 | 0.85 | 0.85 | 0.92 | 0.90 | 0.13 | 0.18 |
| Dou et al.[14] | 0.88 | 0.87 | 0.86 | 0.86 | 0.93 | 0.91 | 0.11 | 0.17 |
| Liu et al.[15] | 0.89 | 0.88 | 0.87 | 0.87 | 0.94 | 0.92 | 0.09 | 0.14 |
| Tian et al.[16] | 0.90 | 0.89 | 0.88 | 0.88 | 0.95 | 0.93 | 0.08 | 0.16 |
| Chen et al.[17] | 0.91 | 0.90 | 0.89 | 0.89 | 0.95 | 0.94 | 0.05 | 0.12 |
| Tong et al.[18] | 0.88 | 0.87 | 0.86 | 0.86 | 0.93 | 0.92 | 0.08 | 0.10 |
| Qian et al.[19] | 0.89 | 0.88 | 0.87 | 0.87 | 0.94 | 0.92 | 0.06 | 0.08 |
| Ours | 0.93 | 0.92 | 0.91 | 0.91 | 0.97 | 0.96 | 0.02 | 0.07 |

The overall comparison shows that the baseline methods perform similarly in terms of discrimination metrics, indicating that conventional features and graph structure modeling can capture some fraud patterns, but there are still limitations in scenarios with complex relationship dependencies and weak signals. The proposed method maintains a leading position in the main classification metrics and demonstrates stronger discriminative power in overall ranking metrics, indicating that the joint modeling of graph representation learning and risk discrimination can more fully utilize the group associations and chain behaviors in the transaction network, thereby improving the ability to identify difficult cases and hidden frauds and reducing the risk of missed detections.

Regarding the quality of risk characterization, the proposed method also outperforms in calibration-related metrics, demonstrating more stable, reliable, and consistent probability outputs with real risks. This phenomenon typically stems from the synergistic effect of two aspects: firstly, neighborhood aggregation and structural regularization suppress noisy relationships, making the representation space smoother and less susceptible to being pulled by local anomalous connections; secondly, loss design and imbalance handling improve the learning quality of scarce positive class samples, allowing the model to maintain discriminative power while reducing overconfidence or unnecessary fluctuations, thus better adapting to the requirements of interpretable decision thresholds and stable alarms in risk control scenarios.

The hidden dimension determines the capacity and information compression of the graph representation, and directly affects the structural semantic strength received by the risk discriminator. Too small a hidden dimension can lead to underfitting of the representation, making it difficult to characterize fine-grained relationship patterns in the transaction network; too large a hidden dimension may introduce redundancy and amplify noise propagation. To examine the impact of variations in representation capacity on risk ranking ability, different hidden dimensions were set, and the corresponding AUROC change trends were recorded. The experimental results are shown in Figure 3.

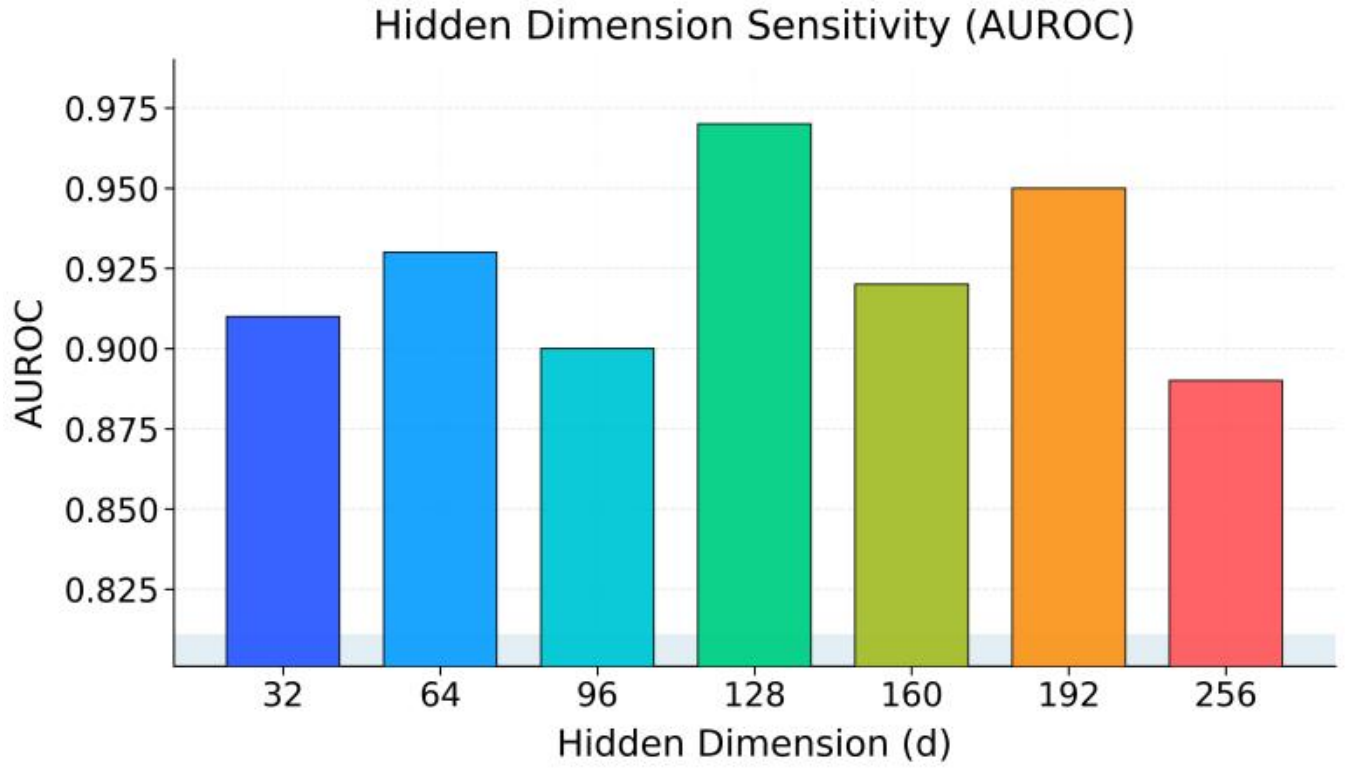

**Figure 3. Sensitivity experiment of hidden dimensions to AUROC**

The change in hidden dimension reveals a clear non-monotonic trend, indicating a strong correlation between representation capacity and risk ranking ability. When the dimension is small, graph structure and attribute information are more easily compressed into coarse-grained representations during multi-layer propagation, making it difficult to characterize the detailed differences in the trading network. As the dimension increases, inter-node relationship patterns and local context are more fully encoded, and the risk score's ability to distinguish similar transactions is enhanced.

When the dimension continues to increase, performance improvement becomes unstable, reflecting that excessively high capacity may amplify the effects of noisy edges and weakly correlated features, making the model more prone to learning accidental co-occurrences rather than generalizable structural signals. This phenomenon is consistent with the sparsity and heterogeneity of financial trading graphs; an overly complex representation space increases unnecessary degrees of freedom and reduces robustness. Therefore, a more suitable range of hidden dimensions needs to be chosen between expressive power and noise resistance stability.

The dropout ratio controls the intensity of random deactivation during representation learning and affects the redundancy suppression and noise resistance of graph structure information during inter-layer propagation. Lower dropout ratios tend to cause the model to overfit to local neighborhood noise, while higher dropout ratios may weaken the accumulation and expression of effective relational signals. To characterize the impact of regularization intensity variations on risk ranking ability, different dropout ratios were set and the corresponding AUROC variation trends were recorded. The experimental results are shown in Figure 4.

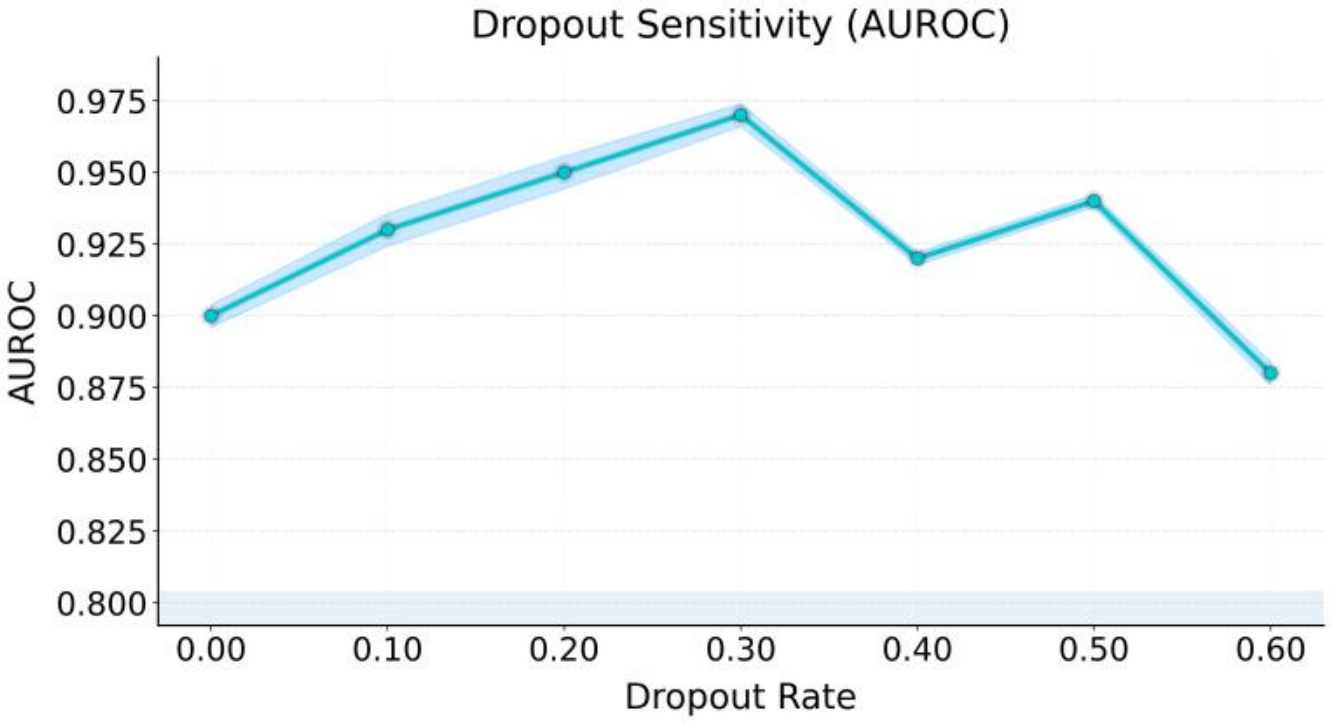

**Figure 4. Sensitivity experiment of dropout ratio to AUROC**

Figure 4 shows a pattern of initial increase followed by a decrease, indicating that Dropout, as a regularization term, strikes a clear balance between suppressing noise propagation and preserving effective structural signals. At lower Dropout strengths, the model more easily remembers accidental co-occurrences and weakly correlated connections in local neighborhoods, making risk ranking more sensitive to noisy edges. As Dropout increases, redundant representations are weakened, structural semantics become more focused, and thus, ranking ability gradually improves and reaches a more ideal range.

When Dropout continues to increase, performance declines with more pronounced fluctuations, reflecting that excessively strong random deactivation weakens the accumulation of key relationship patterns during message passing, leading to unstable representations and difficulty in forming a consistent risk profile. This phenomenon aligns with the sparsity and heterogeneity of financial transaction graphs; excessive regularization can erase the small amount of key neighbor information that should be retained, thus weakening the model's ability to distinguish hidden fraud. Therefore, a moderate Dropout strength is needed to balance robustness and expressive power.

# 5 Limitations

While graph neural networks can effectively utilize transaction relationship information, model performance still depends on the quality of graph construction and the stability of available fields. In real-world business scenarios, key relationship fields may become ineffective due to anonymization, missing information, or strategic changes, leading to sparse edge connections or noise clusters, thus affecting the effectiveness of message delivery and the interpretability of risk scores. Furthermore, relationship rules are typically based on limited prior settings, making it difficult to fully cover more covert fraudulent organizational forms, such as cross-channel and multi-account collaboration. When faced with novel fraud strategies or unseen relationship combinations, representation learning may still introduce biases and reduce generalization ability.

On the other hand, anti-fraud tasks commonly suffer from label lag, extreme class imbalance, and concept drift. Supervised learning frameworks are highly sensitive to label quality and temporal consistency. Even with reweighting and regularization constraints, models may still experience threshold mismatch and probability calibration degradation under long-term distribution changes, affecting the stability of risk level classification and alerting strategies. Meanwhile, graph models have relatively higher training and inference overhead on large-scale transaction networks. Online deployment often requires additional graph sampling, caching, and incremental update mechanisms to meet latency and throughput constraints. These engineering factors also limit the direct migration and consistency of methods across different business systems.

# 6 Conclusion

This paper addresses the real-world needs of financial transaction anti-fraud scenarios, characterized by complex relationship structures, concealed behavioral patterns, and rapid evolution. It proposes a representation learning and risk discrimination modeling approach oriented towards transaction graphs. By unifying multi-source attribute information, such as transactions and identities, into a graph structure and learning node-level risk representations within the message passing framework of graph neural networks, the model can simultaneously utilize individual attributes and neighborhood relationship patterns to form a more structurally semantic risk profile. This provides a more consistent and usable decision-making basis for transaction-level fraud identification. This research emphasizes the synergistic improvement of risk identification capabilities from both the data organization method and the representation learning mechanism, aligning with the core requirements of risk control systems for robustness and scalability.

Methodologically, transaction graph modeling shifts the anti-fraud task from isolated sample discrimination to relationship-driven structured discrimination, helping to explicitly capture key patterns such as homogeneous transaction groups, chain transfers, and collaborative behaviors. The risk discrimination module outputs continuous risk scores based on graph embedding, supporting threshold-based handling and hierarchical management. This facilitates integration with existing business processes and reduces the costs associated with frequent maintenance of rule systems. The overall framework links structural information, representation learning, and discrimination objectives under unified optimization, enhancing its adaptability to noisy connections and scarce fraud samples, and providing a more stable modeling path for identifying potential risks in complex transaction networks.

From an application perspective, this research provides a feasible technical paradigm for financial institutions to improve fraud interception efficiency and risk governance accuracy in high-throughput transaction environments. More importantly, the interpretable clues of relationships provided by graph representation learning can assist in risk tracing and evidence organization in auditing and compliance scenarios, providing structured support for alarm aggregation, identification of black and gray market groups, and cross-channel joint prevention. For business closed loops, continuous risk scores can be linked with manual review, policy rules, and fund disposal, promoting the upgrade from single-point interception to group governance, and establishing a more reasonable technical foundation for reducing false alarm rates, improving user experience, and ensuring transaction security.

Future work can be further advanced in three directions. Firstly, introducing stronger temporal modeling and incremental learning mechanisms to better adapt to concept drift and the rapid emergence of new fraud strategies, and improving the stability and controllability of online updates. On the other hand, focusing on the requirements of explainability and auditability, the framework strengthens the attribution expression of key relationship links and sources of risk contribution, promotes the synergy and consistency between model output, risk control rules, and manual review logic, and enhances business trust and regulatory acceptability. Finally, it explores more efficient graph sampling, caching, and inference optimization strategies in larger-scale and multi-business domain real-world systems to reduce deployment costs and improve cross-scenario migration capabilities, thereby expanding the practical value of this framework in areas such as payment security, digital banking, cross-border settlement, and internet platform transaction governance.

## References

<bib id="bib1"><number>[1]</number>Lu M, Han Z, Rao S X, et al. Bright-graph neural networks in real-time fraud detection[C]//Proceedings of the 31st ACM international conference on information & knowledge management. 2022: 3342-3351.</bib>
<bib id="bib2"><number>[2]</number>Tian Y, Liu G. Transaction fraud detection via spatial-temporal-aware graph transformer[J]. arXiv preprint arXiv:2307.05121, 2023.</bib>
<bib id="bib3"><number>[3]</number>Kim H, Choi J, Whang J J. Dynamic relation-attentive graph neural networks for fraud detection[C]//2023 IEEE International Conference on Data Mining Workshops (ICDMW). IEEE, 2023: 1092-1096.</bib>
<bib id="bib4"><number>[4]</number>Duan Y, Zhang G, Wang S, et al. Cat-gnn: Enhancing credit card fraud detection via causal temporal graph neural networks[J]. arXiv preprint arXiv:2402.14708, 2024.</bib>
<bib id="bib5"><number>[5]</number>Saldaña-Ulloa D, De Ita Luna G, Marcial-Romero J R. A temporal graph network algorithm for detecting fraudulent transactions on online payment platforms[J]. Algorithms, 2024, 17(12): 552.</bib>
<bib id="bib6"><number>[6]</number>Cao K, Zhao Y, Chen H, et al. Multi-hop relational modeling for credit fraud detection via graph neural networks[C]//Proceedings of the 2025 6th International Conference on Computer Science and Management Technology. 2025: 804-809.</bib>
<bib id="bib7"><number>[7]</number>Long S, Cao K, Liang X, et al. Knowledge graph-driven generative framework for interpretable financial fraud detection[C]//Proceedings of the 2025 5th International Conference on Computational Modeling, Simulation and Data Analysis. 2025: 985-990.</bib>
<bib id="bib8"><number>[8]</number>Ou Y, Huang S, Wang F, et al. Adaptive anomaly detection for non-stationary time-series: A continual learning framework with dynamic distribution monitoring[J]. Preprints, 2025.</bib>
<bib id="bib9"><number>[9]</number>Yan R, Ou Y, Sun S, et al. DualShiftNet: Joint class-imbalance and distribution-shift aware learning for business risk prediction[J]. Preprints, 2026.</bib>
<bib id="bib10"><number>[10]</number>Chen N, Sun S, Wang Y, et al. Few-shot financial fraud detection using meta-learning and large language models[C]//Proceedings of the 2025 6th International Conference on Computer Science and Management Technology. 2025: 822-826.</bib>
<bib id="bib11"><number>[11]</number>Ying R, Liu Q, Wang Y, et al. AI-based causal reasoning over knowledge graphs for data-driven and intervention-oriented enterprise performance analysis[C]//Proceedings of the 2025 6th International Conference on Computer Science and Management Technology. 2025: 770-775.</bib>
<bib id="bib12"><number>[12]</number>Chen H, Lu Y, Wei Y, et al. Causal-LLM: A hybrid framework for automated budgetary variance diagnosis and reasoning[J]. Preprints, 2026.</bib>

<bib id="bib13"><number>[13]</number>Cheng D, Wang X, Zhang Y, et al. Graph neural network for fraud detection via spatial–temporal attention[J]. IEEE Transactions on Knowledge and Data Engineering, 2020, 34(8): 3800–3813.</bib>
<bib id="bib14"><number>[14]</number>Dou Y, Liu Z, Sun L, et al. Enhancing graph neural network–based fraud detectors against camouflaged fraudsters[C]//Proceedings of the 29th ACM international conference on information & knowledge management. 2020: 315–324.</bib>
<bib id="bib15"><number>[15]</number>Liu Y, Ao X, Qin Z, et al. Pick and choose: a GNN–based imbalanced learning approach for fraud detection[C]//Proceedings of the web conference 2021. 2021: 3168–3177.</bib>
<bib id="bib16"><number>[16]</number>Tian Y, Liu G, Wang J, et al. ASA–GNN: Adaptive sampling and aggregation–based graph neural network for transaction fraud detection[J]. IEEE Transactions on Computational Social Systems, 2023, 11(3): 3536–3549.</bib>
<bib id="bib17"><number>[17]</number>Chen J, Chen Q, Jiang F, et al. SCN_GNN: A GNN–based fraud detection algorithm combining strong node and graph topology information[J]. Expert Systems with Applications, 2024, 237: 121643.</bib>
<bib id="bib18"><number>[18]</number>Tong G, Shen J. Financial transaction fraud detector based on imbalance learning and graph neural network[J]. Applied Soft Computing, 2023, 149: 110984.</bib>
<bib id="bib19"><number>[19]</number>Qian J, Tong G. Metapath–guided graph neural networks for financial fraud detection[J]. Computers and Electrical Engineering, 2025, 126: 110428.</bib>